# Augmenting the FedProx Algorithm by Minimizing Convergence

Mr. Anomitra Sarkar and Ms. Lavanya Vajpayee
School of Computer Science and Engineering, VIT Chennai, India

*Abstract*— The Internet of Things (IoT) has experienced significant growth and has become an integral part of various industries. This expansion has given rise to the Industrial IoT (IIoT) initiative, where industries are utilizing IoT technology to enhance communication and connectivity through innovative solutions such as data analytics and cloud computing. However, this widespread adoption of IoT is demanding of algorithms that provide better efficiency for the same training environment, without speed being a factor. In this paper, we present a novel approach called G-Federated Proximity (FedProx). Building upon the existing FedProx technique, our implementation introduces slight modifications to enhance its efficiency and effectiveness. By leveraging FTL, our proposed system aims to improve the accuracy of model obtained after the training dataset with the help of normalization techniques, such that it performs better on real time devices and heterogeneous networks. Our results indicate a significant increase in the throughput, approximately 90% better convergence, compared to existing model's performance.

## I. INTRODUCTION

In the rise of the new Industrial Revolution 4.0 or popularly Industrial Internet of Things, security AI and ML along with Real-Time Data sensing/mining is a new trend. The AI model can be trained on a central server by directly sending the user data to a server, which is easier to implement and fast. Yet on the other hand, this poses the biggest threat that comes This work was supported by the SCOPE dept. of VIT University (Chennai) as contribution to development of creation and formulating ideas regarding Smart Cities with the rising age of computational power of computers, "Cyber Security". Sending the data to the server directly makes it vulnerable to all possible options there are to get hacked and leaks. Therefore, Federated Transfer Learning was introduced which in layman language is that model is trained in the device locally and the model weights and parameters are updated on the server timely.

Federated Transfer learning or FTL is an attractive paradigm for fitting a model to data generated by, and residing on, a network of remote devices. Unfortunately, naively minimizing an aggregate loss in a large network may disproportionately advantage or disadvantage the model performance on some of the devices. For example, although the accuracy may be high on average, there is no accuracy guarantee for individual devices in the network. This is exacerbated by the fact that the data are often heterogeneous in federated networks both in terms of size and distribution, and model performance can thus vary widely. In this work, we therefore ask: Can we devise an efficient federated optimization method to encourage a fairer distribution of the model performance across devices in federated networks?

*"McMahan et al., 2017; Li et al., 2019"* paper on the Federated base learning has started this whole world to tackle two key challenges, which distinguish FTL from traditional distributed optimization: high degrees of real systems and statistical heterogeneity. Over the years, continuous development has been seen in this field, as it provides security and privacy but for the cost of speed. So, it is not only important to provide convex solutions to this so that no scope duality is encountered.

Firstly, for homogeneous networks, an iterative approach called FedAvg is aimed to tackle this problem. Federated averaging (FedAvg) is a communication efficient algorithm for the distributed training with an enormous number of clients. In FedAvg, clients keep their data locally for privacy protection; a central parameter server is used to communicate between clients.

The next paradigm being the important task of reintegrating the heterogeneous networks into this FTL problem, called FedProx. FedProx allows for variable amounts of work to be performed locally across devices and relies on a proximal term to help stabilize the method. We provide the convergence guarantees for FedProx in realistic federated settings under a device dis-similarity assumption, while also accounting for practical issues such as stragglers. It resolves the FTL's systems heterogeneity problem and the statistical heterogeneity problem, it contributes as a guaranteed

This work was supported by the SCOPE dept. of VIT University (Chennai) as contribution to development of creation and formulating ideas regarding Smart Cities convergence and more empirical performance for federated learning in heterogeneous networks.

The problem encountered in FedProx and FedAvg and even our FTL based algorithms like Q-FTL are calculating global losses on the basis of all the values of a subset, whereas half of those values are not even contributing to the models, these local devices produce deviations in data that are not helpful enough to provide model update aggregation and may even provide, maybe slight, but a small margin of error in the regression to lean towards the wrong predictive model. To avoid this, we propose the framework below, and provide results that show enough promise for actual development.

Even considering the results of Tian Li and Anit Kumar Sahu's, *"Federated Optimization in Heterogenous Networks"*, provide a sophisticated base for FedProx in tackle, while the results being approx. 22% better than its predecessor. The model produced has obvious noise while being a stable enough model, which is due to the Stochastic Gradient Descent calculations (SGD), which try to provide concavity but do not produce priority for value that are useful. Hence this showcases the importance of segregation in this base of value selection as a whole distribution, talked about in Ghadimi, S. and Lan, G. Stochastics'," First-*and zeroth-order methods for nonconvex stochastic programming."*

In this paper, the new change in the FedProx algorithm is the addition of Boolean term of sparse change compared to last weights. If the weight regarding a topic is not changed locally over a range of accepted change, due to reasons discuss further later in the article, the data will not be updated on the server, therefore aiming to only updated the data that has changed vastly. Making the behavior of the server regarding a client who is constant is, constant for a time-being, and not letting interference hamper the learning models. Our aim in this article is to discuss and provide the articulation of the need of the major required updates in FTL and provide a new console for the implementation.

## II. RELATED WORKS

Federated Transfer Learning's most important aspect is In this section, we introduce the key ingredients behind recent methods for federated learning, including FedAvg, and then outline our proposed framework, FedProx. Federated learning methods (e.g., McMahan et al., 2017; Smith et al., 2017) are designed to handle multiple devices collecting data and a central server coordinating the global learning objective across the network. In particular, the aim is to minimize allocation of proper weights locally on the client and sending the data of the updated model to the server rather than the data. Therefore, depending on the weights, latency, and output of model change, to make the model faster like the former, A Fair resource allocation methodology has been extensively studied in fields such as network management (Ee & Bajcsy, 2004; Hahne, 1991; Kelly et al., 1998; Neely et al., 2008) and the problem is defined as allocating a scarce shared resource. As a service provider, it is important to improve the quality of service for all users while maintaining overall throughput.

The posed problem in FTL is the initial mode to be trained, algorithms like FedAvg or Federated Average tackle this solution by providing the clients a shared trained model and then updating the model accordingly. But in the assumption of FedAvg doesn't hold good because it promises to optimize data in homogeneous environment settings. Whereas Algorithms like FedProx, which optimizes data in heterogenous environment settings, by penalizing and minimizing loss in a device and providing a better convergence overall.

While in recent years, better algorithms like FedProx and q-FTL have been proposed. FedProx objective is flexible enough to enable trade-offs between fairness and other traditional metrics such as accuracy by changing the parameter/weights of heterogeneity and sparse data.

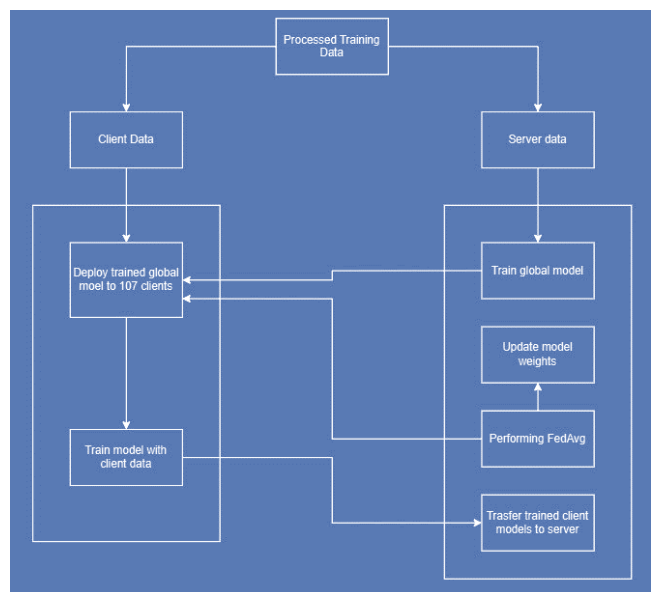

*This figure represents the Working of a FTL Model*

For the second part of the article, regarding additional security parameters just as validating local IOT digital certificates simply don't work. Considering the resource limitations of IIoT devices and the performance overhead introduced by strong security mechanisms, the manufacturers and developers of these devices frequently refuse to implement costly certificate validation mechanisms.

Even more so in the case of brokered communication mechanisms, common in industrial contexts, where communications are short-lived but very frequently established. Therefore, the TLS/DTLS session handshake introduces a high overhead per every transmitted message.

Therefore, a previously introduced innovative in-network certificate validation system along with the FedProx system is used to securely and efficiently train and broadcast IIOT communication along DTLS client and servers, respectively.

The idea proposed is directed to future work as the FTL is growing day by day but there is still no unified architecture for the model implementation. The paper promises to provide not only the theoretical approach of FTL concepts but also the practical aspects of FTL in IIOT communications.

### III. UNDERSTANDING PREVIOUS FTL OPTIMIZATIONS

In this section, we introduce the key ingredients behind recent methods for federated learning, including FedAvg, and then outline our proposed framework, FedProx. Federated learning methods (e.g., McMahan et al., 2017; Smith et al., 2017) are designed to handle multiple devices collecting data and a central server coordinating the global learning objective across the network. In particular, the aim is to minimize:

$$\min f(w) = \sum_{k=1}^{N} p_k F_k(w) = E_k[F_k(w)]$$

where N is the number of devices, $p_k \geq 0$, and $\sum_k p_k = 1$. In general, the local objectives measure the local empirical risk over possibly differing data distributions $D_k$, 7i.e., $F_k(w) := E_{xk} \sim D_k[f_k(w;x_k)]$, with $n_k$ samples available at each device k. Hence, we can set $p_k = (n_k)/n$, where $n=\sum\_k(n_k)$ is the total number of data points. In this work, we consider $F_k(w)$ to be possibly non-convex.

In general, the local objectives measure the local empirical risk over possibly differing data distributions $D_k$, i.e., $F_k(w) := E_{xk} \sim D_k[f_k(w;x_k)]$, with $n_k$ samples available at each device

To reduce communication, a common technique in federated optimization is that on each device, a local objective function based on the device's data is used as a surrogate for the global objective function. At each outer iteration, a subset of the devices is selected, and local solvers are used to optimize the local objective functions on each of the selected devices. The devices then communicate their local model updates to the central server, which aggregates them and updates the global model accordingly. The key to allowing flexible performance in this scenario is that each of the local objectives can be solved inexactly. This allows the amount of local computation vs. communication to be tuned based on the number of local iterations that are performed (with additional local iterations corresponding to more exact local solutions). We introduce this notion formally below, as it will be utilized throughout the paper.

***Definition 1** (β-inexact solution).* For a function $h(w,w_o) = F(w) + \frac{\mu}{2}||w - w_o||^2$, and $\beta \in [0,1]$, we say $w^*$ is a β-inexact solution of $\min_w f(w;w_o)$ if $||\nabla h(w^*;w_o)|| \leq \beta||\nabla h(w_o;w_o)||$, where $\nabla h(w_o;w_o) = \nabla F(w) + \mu(w-w_o)$. Note that a smaller β corresponds to higher accuracy.

*(Refer FTL for Heterogeneous Data, 2022)

### 3.1 Federated Averaging (FedAvg)

**Algorithm 1 (FedAvg)**

Input: K, T,$\eta$,E,$w^o$, N, $p_k$, k=1,..., N
for t = 0,…,t-1 do
    Server selects a subset $S_t$ of K devices at random
    (each device k is chosen with probability $p_k$)
    Server sends $w^t$ to all chosen devices.
    Each device k ∈ $S_t$ updates $w^t$ for E epochs of SGD
    on $F_k$ with step size $\eta$ to obtain $w_k^{t+1}$
end for

In Federated Averaging (FedAvg)(McMahan et al.,2017), the local surrogate of the global objective at device k is $F_k$ (.), and the local solver of stochastic gradient descent(SGD), with the same learning rate and number of local epochs used on each device. At each round, a subset K << N of the total devices are selected and run SGD locally for E numbers of Epochs, and then the resulting model updates are averaged. The details of FedAvg are summarized in algorithm 1.

$$\min f(w) = = \sum_{k=1}^{N} p_k F_k(w) = E_k[F_k(w)]$$

### 3.2 Federated Proximity (FedProx)

FedProx (Algorithm 2), is like FedAvg in that a subset of devices are selected at each round, local updates are performed, and these updates are then averaged to form a global update. However, FedProx makes the following simple yet critical modifications, which result in significant empirical improvements and allow us to provide convergence guarantees for the method. Tolerating partial work. As previously discussed, different devices in federated networks often have different resource constraints in terms of the computing hardware, network connections, and battery levels. Therefore, it is unrealistic to force each device to perform a uniform amount of work (i.e., running the same number of local epochs, E), as in FedAvg. In FedProx, we generalize FedAvg by allowing for variable amounts of work to be performed locally across devices based on their available systems resources, and then aggregate the partial solutions sent from the stragglers (as compared to dropping these devices).

***Definition 2*** ($\beta_k^t$ - inexact solution). For a function $h(w,w_o) = F_k(w) + \frac{\mu}{2}||w-w_o||^2$, and
$\beta \in [0,1]$, we say w* is a $\beta_k^t$ - inexact solution of $\min_w h_k(w;w_t)$ if
$||\nabla h_k(w^*;w_t)|| \leq \beta_k^t||\nabla h(w_t;w_t)||$, where $\nabla h_k(w;w_t) = \nabla F_k(w) + \mu(w-w_t)$. Note that a smaller $\beta_k^t$ corresponds to higher accuracy.

Analogous to Definition 1, $\beta_k^t$ measures how much local computation is performed to solve the local subproblem on device *k* at the *t*-th round. The variable number of local iterations can be viewed as a proxy of $\beta_k^t$. Utilizing the more flexible $\beta_k^t$-inexactness, we can readily extend the convergence results under Definition 1 to consider issues related to systems heterogeneity such as stragglers.

**Proximal term:** As mentioned in above, while tolerating nonuniform amounts of work to be performed across devices can help alleviate negative impacts of systems heterogeneity, too many local updates may still (potentially) cause the methods to diverge due to the underlying heterogeneous data.

$$\min h_k(w,w^t) = F_k(w) + \frac{\mu}{2}||w-w^t||^2$$

The proximal term is beneficial in two aspects: (1) It addresses the issue of statistical heterogeneity by restricting the local updates to be closer to the initial (global) model without any need to manually set the number of local epochs. (2) It allows for safely incorporating variable amounts of local work resulting from systems heterogeneity. We summarize the steps of FedProx in Algorithm 2.

| **Algorithm 2 FedProx(Base Framework)** |
|---|
| Input: K, T, $\mu$, $\beta$, $w^o$, N, $p_k$, k=1,..., N |
| for t = 0,...,T-1 do |
|     Server selects a subset $S_t$ of K devices at random (each device k is chosen with probability $p_k$) |
|     Server sends $w^t$ to all chosen devices. |
|     Each chosen device k $\in S_t$ finds a $w_k^{t+1}$ which is a $\beta_k^t$ - inexact minimizer of: |
|     $w_k^{t+1} \approx \arg\min_w h_k(w;w^t) = F_k(w) + \frac{\mu}{2}||w-w^t||^2$ |
|     Each device k $\in S_t$ sends $w_k^{t+1}$ back to the server. |
|     Server aggregates the w's as $w^{t+1} = \frac{1}{k}\sum_{k \in s} w_k^t w_k$ |
| end for |

## IV. PROPOSED FTL OPTIMIZATION

### 4.1 Architecture

In this section, we propose an optimization methodology for the same FTL convergence for heterogeneous and sparse data. This is very similar to the previous FedProx algorithm with addition of a Function $G(w^t)$ such that,

$$\min h_k(w,w^t) = F_k(w^t) + G(w^t) \mu ||w-w^t||^2$$

where,

$$G(w^t) = \begin{cases} \frac{f(w^t,\mu_s,\sigma_s)}{2}, & \mu - 2\sigma < w < \mu + 2\sigma \\ 0, & Otherwise, \end{cases}$$

$\mu_s$=mean of server's mean
$\sigma_s$=standard deviation of server
(Rest hold their respective meanings)

This Scalar Multiplication of function with the proximity term ensures the following:
- Local client machine whose data are sparser and out of order hence providing error on global model, will be rejected and values that are closer to the accepted deviation rate of normal distribution (>95%) of data epochs are validated for convergence calculations and providing final updates to the model.
- Creating a better convergence of regression model even for more realistic machines, the indirect epochs negations let the global model preserve the actual expected result without getting disturbed due to low performing client models.

In cumulation, adding a Filter Boolean Function/Term G that disposes any client data set at any $t^{th}$ iteration that was obtained and preprocessed to get a distributive epoch outside the acceptance area of $w_t \in (\mu_S - 2\sigma_S, \mu_S + 2\sigma_S)$ or more than 95% deviation from expected mean / first moment of distribution derived from global server regression models and calculating the model accordingly.

To let the heterogeneity and sparse device with more faulty data with latency factors too contribute to ML training while not affecting the global server's convergence model. Whilst provide an overall better output which are closer to $\mu_s$ (or mean expectation of server's regression model)

### 4.2 Why Gaussian Normalization

In a normal distribution, data is symmetrically distributed with no skew. When plotted on a graph, the data follows a bell shape, with most values clustering around a central region and tapering off as they go further away from the center. All kinds of variables in natural and social sciences are normally or approximately normally distributed. Height, birth weight, reading ability, job satisfaction, or SAT scores are just a few examples of such variables.

$$f_n(w^t,\mu_s,\sigma_s) = \frac{1}{\sigma_s\sqrt{2\pi}} e^{-(w^t-\mu_s)^2} \frac{1}{e^{2\sigma_s^2}}$$

Considering the data set is U and after the pre-processing and data augmentation U => V. The data set is cleaned and multiplied to provide for the spaces and create abundance of training datasets. Techniques like rotation, inversion and scaling are often used when dealing with image recognition machine learning data augmentation.

It is show in this exponential growth of data and further calculations any natural regression parameter at any $t^{th}$ epoch will tend to fall into the *Standard Gaussian Normal Distribution*, therefore recent studies regarding Maximum Likelihood Estimation of Parameters using Gaussian Distribution to estimate probability and statistical correlations of the data with regards to the first $r^{th}$ moment or the expectation of the data set, the mean, and the standard deviation. The mean is the location parameter while the standard deviation is the scale parameter.

$$z = \frac{x_i - \mu}{\sigma}$$

*(z values are the distribution values derived from the standard normal distribution)

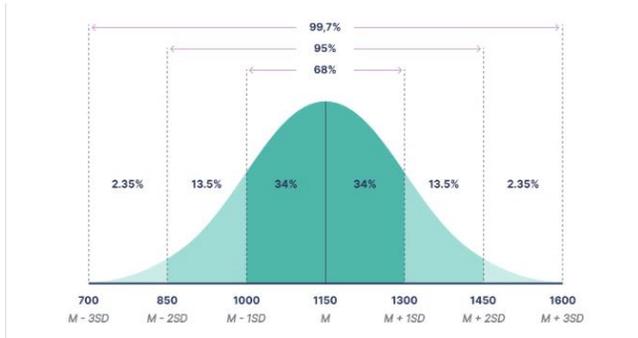

The above is a general normal distribution for mean M and standard deviation SD in 1D distribution. For our parameter, we are proposing to only conclude the inclusion of 95% of data points that are closer to global server mean and reject the rest of the points.

This was the only disadvantage of FedProx calculating the convergence term for all values that are out of bounds to even satisfy the cubic interpolation and kurtosis accordingly. Therefore, here we only consider the $w^t$ loss weight if and only if it's within 2 times the standard deviation region.

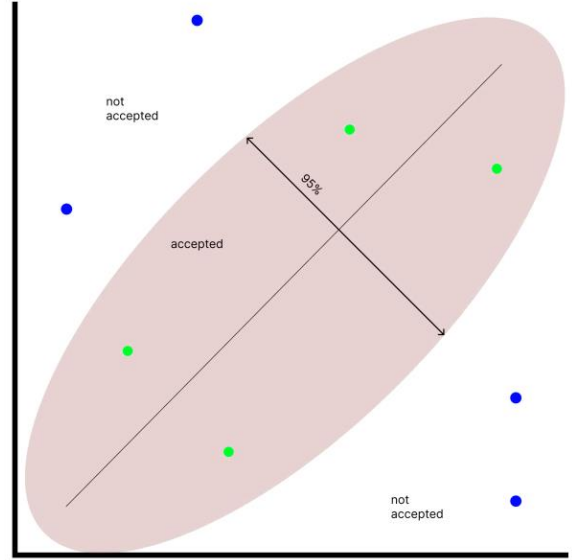

The above is a 2D distribution showcase with respect to a sample regression line and the normal distribution and acceptance region implemented in a 2D space. In the next article, we discuss the correctness of our argument via proving that if all the weights were considered then the proposed architecture would result in the general FedProx system.

So the $f_n(w^t, \mu_s, \sigma_s)$ is a normalization function that follows:

$$f_n(w^t, \mu, \sigma) = \frac{1}{\sigma\sqrt{2\pi}} e^{-(w^t - \mu)^2} \frac{1}{e^{2\sigma^2}}$$

where, $\mu_s = \mu$ and $\sigma_s = \sigma$

### 4.3 Proof of Equivalence

If considering previous FedProx model where all the values of $w^t$ are allowed for convergence-minimization calculation for the final model updates, to check for verification, we also must consider the cumulative of all values.

$$\min h_k(w, w^t) = F_k(w^t) + G(w^t) \mu \|w - w^t\|^2$$

where,
$$G(w^t) = \begin{cases} \frac{f(w^t, \mu_s, \sigma_s)}{2}, & \mu - 2\sigma < w < \mu + 2\sigma \\ 0, & Otherwise, \end{cases}$$

Therefore, here to prove equivalence to previous FedProx we need to accept 100% of values instead of 95% values at a difference double standard deviation from the mean. So, the second condition of Nullifying the function for other values is removed.

$$G(w^t) = \begin{cases} \frac{f(w^t, \mu_s, \sigma_s)}{2}, & \mu - 2\sigma < w < \mu + 2\sigma \\ \underline{\phantom{xx}} & \end{cases}$$

Now that all values are calculated for weights, applying gaussian normalization to all factors, sum all the values, in the standard normalization function "$f_n$"

Sum of $f_n$ is a Cumulative mass/density function for all $i^{th}$ level iterations of servers till n (all values).

$$\sum_{i=0}^{n} G(w) = \sum_{i=0}^{n} f(w^t, \mu_s, \sigma_s)/2$$

We know that in both discrete or continuous statistical distribution, the sum of all probabilities is always equal to 1. (Property of sure event)

$$\sum_{i=0}^{n} P(x) = 1 \text{ or } \int_{-\infty}^{\infty} P(x)dx = 1$$

So, the equation 1 becomes

$$\min h_k(w,w^t) = F_k(w) + \frac{\mu}{2}\|w-w^t\|^2$$

as $f_n$ becomes 1

Hence, we obtain the original FedProx Equation and hence prove the correctness and equivalence of choosing a standard normal distribution and corresponding correlation with the FedProx model, to only provide a slight betterment in overall performance in the proposed solution.

**4.4 Role of 5% Significance**

In statistics, before rejecting any value out of the dataset, it is required to be sure that rejected value does not affect the statistical model values by a considerably large value such that it still instills the quality of the previous model only.

One might ask why 5% of values from the value distribution is rejected instead of some other random percentage selection, it is because it depends on the level of significance. The **level of significance**, also known as alpha (α), is a critical concept in statistics that measures the strength of evidence required to reject the null hypothesis and conclude that an effect is statistically significant. It is determined by the researcher before conducting the experiment and represents the probability of rejecting the null hypothesis when it is true. For example, a significance level of 0.05 indicates a 5% risk of concluding that a difference exists when there is no actual difference.

In statistical hypothesis testing, if the p-value of an observed effect is less than or equal to the significance level, an investigator may conclude that the effect reflects the characteristics of the whole population, thereby rejecting the null hypothesis. The significance level is essentially a measure of how confident one can be about rejecting the null hypothesis and plays a significant role in determining the strength of the evidence required to make statistical inference. There exist only three levels of significance: 1%, 5% and 10%.

The 5% level of significance is considered standard in statistics for several reasons. One key factor is historical convention, as it has been widely used for many years. Additionally, a significance level of 0.05 indicates a 5% risk of concluding that a difference exists when there is no actual difference. Lower significance levels indicate that stronger evidence is required before rejecting the null hypothesis. This threshold is a balance between the need for strong evidence and the practicality of obtaining such evidence in various research settings. It also aligns with the balance between the risk of making a Type I error (incorrectly rejecting a true null hypothesis) and the need to detect meaningful effects. Therefore, the 5% level of significance has become a widely accepted standard in statistical analysis due to its historical usage, practicality, and the balance it strikes between the risk of errors and the need for strong evidence.

Our choice 5% is very intricate and important for level of significance of model disturbance avoided and model relevance preserved. Firstly, logically if considered 1% level of significance, accordingly it is provided that only a right tailed and left tailed distribution is provided for this level of significance, which will result in heterogeneity in acceptance and deviate the model, as the data and device is heterogeneous, the data accepted should in homogeneous in nature, or it will shift the nature of the distribution to the opposite of rejection region.

Whereas 5% and 10% level of significance has two tailed rejection regions have a upper hand as they both reject from both sides of distribution, hence not disturbing the model. But the main reason for not considering 10% to leave scope for updating the model whilst providing a better scope for model training and not accepting values, which in turn are not adhering to the standard deviation limits. The 5% level of significance for two tailed distribution rejection overcomes the former and the later as the 95% in a gaussian normal distribution is equivalent to the factored limits of 2 times the standard deviation from about the mean of the model making calculation and further analysis easier.

Moreover, 5% level of significance is the globally attested level of significance and perceived as the default for finding and comparing the test statistics of Null hypothesis.

**4.5 G-Kernel**

In the proposed framework, the final addition will be creating a separate console for our architecture, making it a parallel running process makes it better for adding onto already existing models. The G-Kernel will consist of two outputs, the current calculated parameter calculated according to the methodology provided above, the second output is a performance calculator, "$n_0$". This parameter is calculated by no of epoch losses that were rejected by the G-Kernel, and then dividing the result to the total sample taken into consideration. The result of $n_0$ varies from [0,1] for 0, it's a ideal performing device and for 1 it is completely faulty model.

This helps in the identification of local client device performance and correct it for future reference. Also, this helps us in deriving a kurtosis base i.e. skewness of the distribution obtained from the model. The relationship is as follows,

$$3n_0 \; \alpha \; k$$

The $n_0$ is directly related to the skewness of the graph, which might help in deriving important statistical correlations regarding the model, where k can be calculated as,

$$k = \frac{\mu_{s4}}{\sigma_{s4}} = \frac{n(n+1)}{(n-1)(n-2)(n-3)} \frac{\sum(w_t - \mu_s)^4}{(\sum(w_t - \mu_s)^2)^2} - 3\frac{(n-1)^2}{(n-2)(n-3)}$$

The Kurtosis results as follows,

| Type of Kurtosis | Tailedness | Outlier Frequency | Kurtosis Value |
|---|---|---|---|
| *Leptokurtic* | Heavy | High | >3 |
| *Mesokurtic* | Moderate | Moderate | 3 |
| *Platykurtic* | Light | Low | <3 |

## V. EXPERIMENTS

We generalize the simulations and present the observations and empirical results regarding the Proposed framework, we demonstrate the improved performance of the G-Kernel Implementation by converging the data more towards the observed server mean, regression data. In the next section we discuss the derivation of the dataset for the experiment and how the convergence results are obtained for the provided dataset. The next section also discusses the settings for the model experiment and afterwards we explain the derivation of our observations accordingly. We provide all the code, data and experiments are publicly available at: https://github.com/AnomitraSarkar/G-FTL-Optimization

**5.1 Experiment Details**

For the given experiment we observe the effects on the results obtained for the general data parameters which are the mean of data on any two attributes, for our datasets its "Height" and "Weight". For the dataset it is derived from Kaggle and is a general Height vs Weight Dataset for about 25000 observations, loaded into a CSV (comma separated values) file, namely, "SOCR-HeightWeight.csv". Where the SOCR is Statistics Online Computational Resource Data, by Smit Patel updated 4 years ago. In words of Smit Patel the dataset is described as follows:

This is a simple dataset to start with. It contains only the height (inches) and weights (pounds) of 25,000 different humans of 18 years of age. This dataset can be used to build a model that can predict the heights or weights of a human. Scraped the derived dataset from aforementioned HTML page using a Regex Parser. *[19] links to HTML data page.

The experiments are conducted in Python, using mathematical and statistical library: MatPlotLib's PyPlot. The experiment is divided into three parts. Firstly, analyzing the raw data and plotting a regression line, to obtain the expected mean from normal linear regression methodology. Secondly, we filter the data on the boundary conditions, monitoring the actual mean and standard deviation derived from the dataset, we filter the data on the 95% significance limits. Hence, we again plot the data and plot the regression line, to obtain the expected mean from optimized methodology, again using the linear regression technique. Thirdly, drawing the confidence in the results, we compare the deviations of predicted mean from both techniques to the actual mean of the datasets, and deviations from each other are also calculated.

This experiment is only conducted on a local device of Windows 11 x64 architecture and Python version 3.12.0 and Matplotlib library version 3.8. The experiment showcases no fancy calculations like K-neighboring cluster or any other machine learning algorithm on the namely dataset. It simply focuses on the most popular statistical parameter for a dataset to derive the conclusions and make the respective point clearly, showcasing the accuracy of the proposed technique and inferring the overall improvement of values obtained for the model compared to the normal.

**5.3 Implementation**

Initially the first task is visualizing the raw data, and deriving the actual data parameters, and then using a linear regression model to get the expected data parameters. The below diagram shows the data plotted Height vs Weights and the red line shows the linear regression line that maps exactly middle from minimum of the dataset to the maximum of the dataset.

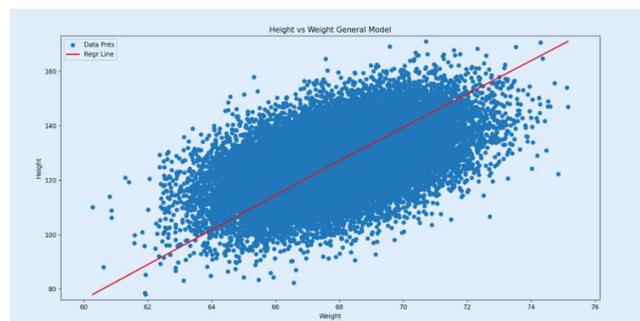

Now moving onto the data obtained we will be discussed, in the results section and plotting the filtered data which is bounded on the $\mu - 2\sigma < w < \mu + 2\sigma$ on both dataset value and applying the same linear regression technique to derive the same as above. On filtering the data, the shape of the data reduces as follows, (25000,0) (25000,0) to (23865,3) (23821, 3) and then utilizing this immediately for recording the observation and the concurrent graph showcases the discussed, in the same way.

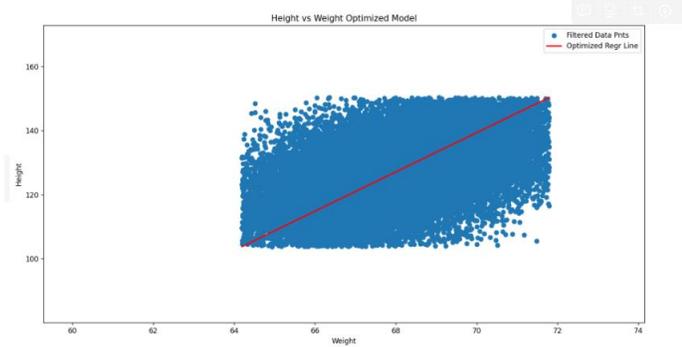

The difference between the two graph is obvious and it is easy to derive conclusions based on visual aspects, but the experiment depends on the obtained convergence on predicted mean, mathematically, that points the actual convergence of the client model with respect to proposed updates in server model.

**5.4 Results**

Firstly, depicting the data parameters from the datasets and using the linear regression on normal dataset and optimized data accordingly.

| OBSERVATION | HEIGHT | WEIGHT |
|---|---|---|
| ACTUAL | 67.9931 | 127.079 |
| NORMAL | 67.7156 | 124.469 |
| OPTIMIZED | 67.9929 | 127.073 |

It is very important to note that for both the values, normal and optimized techniques, result in the same value till the 10's place value but differs in the decimal part of the result. We further analysis the deviations to get the better insights, the deviations are as follows:

| DEVIATIONS | HEIGHT | WEIGHT |
|---|---|---|
| NTA | 0.408179 | 2.05387 |
| OTA | 0.000270023 | ~0 |
| OTN | 0.409581 | 2.09141 |

As for the inference derived from the above table, we can say the deviations obtained from optimized with respect to actual value are far lesser than the deviations obtained from the normal technique with respect to actual value. The optimized technique produced an error of nearly 0 while the normal technique gives a significant enough error. This signifies that for deriving the correct and better models, that approach the actual data parameters, dataset values that lie in the server's 95% data distribution are far better than considering all other dataset values. As they deviate the data parameters from their actual expected value, and provide a far inefficient convergence of model performance, towards the client side.

The percentage value that signifies the amount of decrease in deviations or percentage increase in accuracy with respect to deviations from the actual values, are as follows:

| RESULT | VALUE (%) |
|---|---|
| HEIGHT | 99.9338 |
| WEIGHT | 99.7367 |

Here,
NTA: Normal To Actual
OTA: Optimized To Actual
OTN: Optimized To Normal

The deviations from mean have been reduced more than 99% which means the optimized prediction compared to the normal prediction will be 99% closer, relatively. The above result means the deviation from optimized method is closer by 99% compared to the deviation from actual mean of regression model, found via normal method.

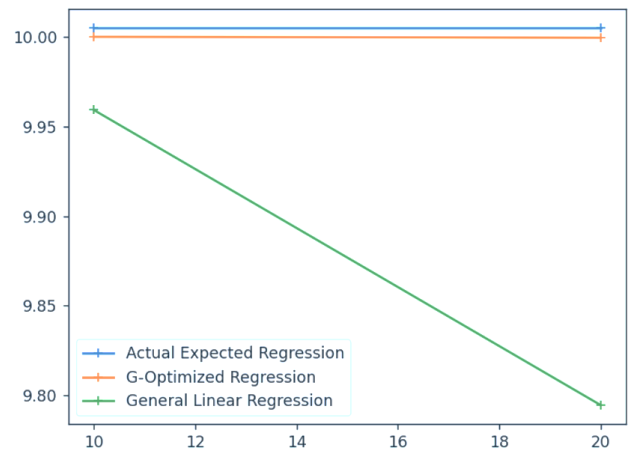

The above showcased Actual Expected Regression line is provided after introducing an error of 0.005 on normalized value of every regression value on it, this is to showcase the close approximation of each value obtained as on the G-optimized regression and negate a little overlapping of the model since the error produced is near to negligible.

As for the calculations and formulae bases used to obtain the percentage values, Formula used is (x100%):

$$\frac{OTA - NTA}{NTA} = \frac{OTN}{NTA}$$

Repeating the same for *Iris* data set a very popular in the machine learning modelling environment and is the most important beginner dataset that is introduced for predication on labelling-based models. Filter the same way as done above we get the following results for the properties namely "sepal width vs petal length". The observations are:

| OBSERVATION | SEPAL WIDTH | PETAL LENGTH |
|---|---|---|
| ACTUAL | 3.05733 | 3.758 |
| NORMAL | 3.2 | 3.95 |
| OPTIMIZED | 3.05 | 3.95 |

The above observations give out the deviation matrix that is as follows, attributes do follow the same naming scheme, we get:

| DEVIATIONS | SEPAL WIDTH | PETAL LENGTH |
|---|---|---|
| NTA | -4.66638 | 2.05387 |
| OTA | 0.23986 | ~0 |
| OTN | -4.6875 | ~0 |

The results on the *Iris* dataset to concur the efficiency of our model for base dataset and on the mentioned properties are the following:

| RESULT | VALUE (%) |
|---|---|
| SEPAL WIDTH | 105.14 |
| PETAL LENGTH | ~0 |

Plotting the above for drawing inferences that showcases that the optimized regression after the filter parameters provided closer approximation compared to the former.

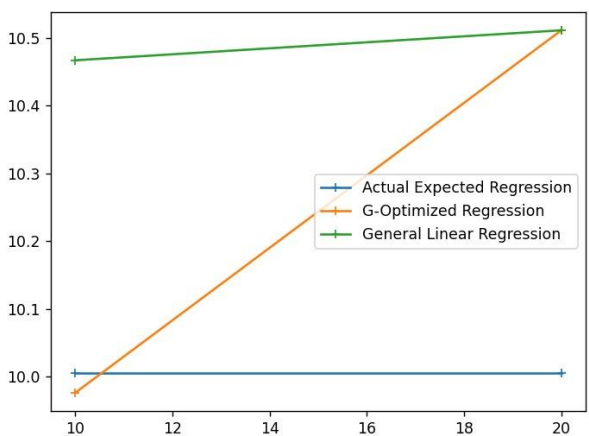

Concluding the result with one more medical dataset, to showcase the attributes real-life based applications and helping in providing better approximation which is in fields like medicine example organ segmentation, disease identification and tumor detection, etc.

Taking it a step further and implementing the same filter and normalization for the *Heart Diseases Stats from the Hungary Cleveland* dataset while monitoring the "sex vs chest pain type" fields which come under the general perspective classes. The results are the following.

| OBSERVATION | SEX | CHEST PAIN TYPE |
|---|---|---|
| ACTUAL | 0.76 | 3.23277 |
| NORMAL | 0.5 | 2.5 |
| OPTIMIZED | 0.5 | 3 |

For the above observations the deviation matrix obtained is as follows:

| DEVIATIONS | SEX | CHEST PAIN TYPE |
|---|---|---|
| NTA | 34.5435 | 22.667 |
| OTA | 34.5435 | ~0 |
| OTN | ~0 | 20 |

The results on the *Heart disease final stat log* dataset to concur the efficiency of our model for medicinal based dataset and on the mentioned properties are the following:

| RESULT | VALUE (%) |
|---|---|
| SEX | ~0 |
| CHEST PAIN TYPE | 68.2339 |

Making the concluding plot the again showcases that the Optimized Regression is a much more closer approximation that the General Linear Regression.

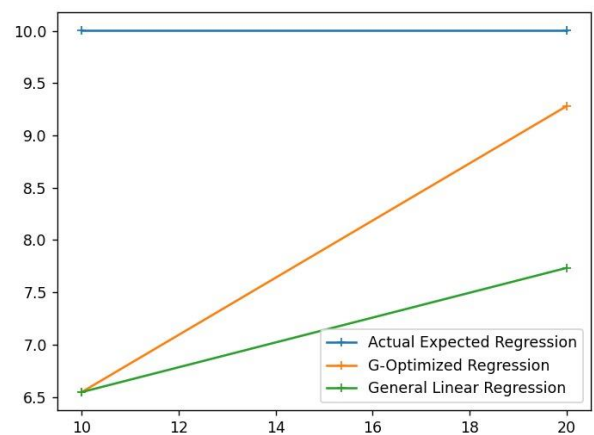

## VI. CONCLUSION

**6.1 Proposed Architecture**

The cumulation of the above discussed framework, irrespective of the G-Kernel is given as the final algorithm is:

---
**Final Algorithm (Proposed Framework)**

**Input:** K, T, μ, β, $w^o$, N, $p_k$, k=1,..., N
**for** t = 0,…,T-1 **do**
    Server selects a subset $S_t$ of K devices at random (each device k is chosen with probability $p_k$)
    Server sends $w^t$ to all chosen devices.
    Each chosen device k ∈ $S_t$ finds a $w_k^{t+1}$ which is a $β_k^t$ - inexact minimizer of:
    $w_k^{t+1} \approx \arg\min_w h_k(w;w^t) = F_k(w) + G(w^t) \mu \|w-w^t\|^2$
    (G-Kernel):
    If for $w^t \in [\mu - 2\sigma, \mu + 2\sigma]$,
        Then do, return normal value according to provided function.
    Else
        Then do, return G = 0
        Also, $N_0=N_0+1$
    Each device k ∈ $S_t$ sends $w_k^{t+1}$ back to the server.
    Loopback to Client till "t" epochs are done, monitor $N_0$
        $N_0$ = Count [$w^t$ rejected as $G(w^t) == 0$ )
    If Epochs Completed, $N_0 = N_0/N$ provide to one terminal
    Server aggregates the w's as $w^{t+1} = \frac{1}{k}\sum_{k \in s} w_k^t w_k$

**end for**

---

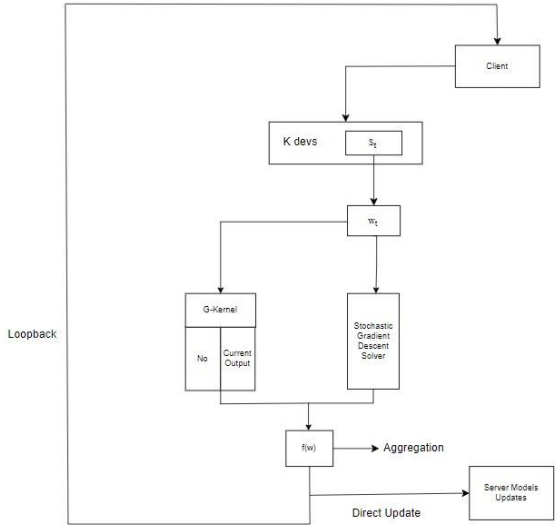

The above diagram represents the cumulation of our idea in its entirety. It showcases the flow of a general FTL model and its complete workflow integration with the G-Kernel which is a console base program that servers two goals, firstly, to calculate the G Parameter and feeding it two the model while the convergence is calculated parallely. Secondly to calculate the n0 value which measures the relevance or actual accuracy of the model, and it ranges from 0 to 1. if no = 0, then the device is a good contributor and is the ideal device for model updates. if no = 1, then the device is a faulty device and may need alteration or corrections in the future.

**6.2 Future Works**

The above proposed model architecture has two aspects that make it optimal for real world implications in the future.

- The G-Kernel is a direct parallel console, so it can be directly added onto any current FedProx algorithm as an optimizer that can augment the entire model. Creating a custom parameter for rejection rate, in the future, a certain server can toggle this level of significance depending on the environment.

- The G-Kernel's 2nd output helps us to review the devices that are faultier and are not working up to the mark, this provides us a mechanism to find out certain devices that are not useful at all and may need corrections.

In addition to the above the proposed framework is easy to understand and implement while also improving the efficiency hence the speed of model training and updating it at better cost and convergence. For median level device, 50% accuracy of data weights losses that are not sparse. The increase in efficiency may reach well over 20% with respect to previous model, in terms of convergence obtained on the regression cluster after updating from the client.

Also, the future works may rely heavenly on implementing a similar G-Kernel in the Q-FTL Algorithm to achieve the same and providing the no value from the G-Kernel helps is achieving a better insight in fairness weight allocation in the Q-FTL Algorithms and Neural Networks.

The better efficiency and parallel calculations help in increasing the efficiency at no cost of speed; therefore, the dataset can be increased accordingly without worrying about training model disturbances. hence reducing the margin of error substantially with regards to the previous models. The convergence correction of the regression model and speed of the proposed framework is already ensured by the FedProx Optimization, the proposed framework optimizes this algorithm to its maximum such that it converges to more accurate regression cluster and does not let any hinderance from faulty devices to affect the model.